\documentclass[letterpaper]{article} 
\usepackage{aaai23}  
\usepackage{times}  
\usepackage{helvet}  
\usepackage{courier}  
\usepackage[hyphens]{url}  
\usepackage{graphicx} 
\usepackage{natbib}  
\usepackage{caption} 
\usepackage{algorithm}
\usepackage{algorithmic}

\usepackage{amsmath,amssymb}
\usepackage{amssymb}

\usepackage{booktabs}
\usepackage{pifont}
\newcommand{\cmark}{\ding{51}}%
\usepackage{multirow}

\usepackage{newfloat}
\usepackage{listings}
\DeclareCaptionStyle{ruled}{labelfont=normalfont,labelsep=colon,strut=off} 
\floatstyle{ruled}
\newfloat{listing}{tb}{lst}{}
\floatname{listing}{Listing}
\title{Context-Aware Transformer for 3D Point Cloud Automatic Annotation}
\author{
    Xiaoyan Qian, 
    Chang Liu,
    Xiaojuan Qi,
    Siew-Chong Tan, \\
    Edmund Lam,
    Ngai Wong 
}
\affiliations{
    The University of Hong Kong, Pokfulam, Hong Kong \\

    \{qianxy10, lcon7\}@connect.hku.hk,
    \{xjqi, sctan, elam, nwong\}@eee.hku.hk\\
}

\usepackage{bibentry}

\begin{document}

\maketitle

\begin{abstract}
3D automatic annotation has received increased attention since manually annotating 3D point clouds is laborious. However, existing methods are usually complicated, e.g., pipelined training for 3D foreground/background segmentation, cylindrical object proposals, and point completion. Furthermore, they often overlook the inter-object feature relation that is particularly informative to hard samples for 3D annotation. 
To this end, we propose a simple yet effective end-to-end Context-Aware Transformer (CAT) as an automated 3D-box labeler to generate precise 3D box annotations from 2D boxes, trained with a small number of human annotations. We adopt the general encoder-decoder architecture, where the CAT encoder consists of an intra-object encoder (local) and an inter-object encoder (global), performing self-attention along the sequence and batch dimensions, respectively. The former models intra-object interactions among points, and the latter extracts feature relations among different objects, thus boosting scene-level understanding.
Via local and global encoders, CAT can generate high-quality 3D box annotations with a streamlined workflow, allowing it to outperform existing state-of-the-art by up to 1.79\% 3D AP on the hard task of the KITTI \emph{test} set.

\end{abstract}

\section{Introduction}
\label{sec:introduction}
3D point cloud has emerged as indispensable sensory data in 3D visual tasks, driven by the ubiquity of the LiDAR sensor and its widespread applications in autonomous driving and robotics. Such inherence has led to the rapid development of 3D object detectors~\cite{shi2019pointrcnn,lang2019pointpillars,xu2021paconv} focusing on the problem of identifying and localizing objects in 3D scenes. Nevertheless, these 3D object detectors require large amounts of 3D ground truth for supervision signals. Although 3D data acquisition is accessible by modern LiDAR scanning devices, it is laborious, error-prone, and even infeasible in some extreme real-world applications to manually annotate 3D point clouds for training high-quality 3D object detection models ~\cite{wei2021fgr,qin2020vs3d, MAPGen2022}. The difficulty, therefore, motivates us to develop 3D automatic annotators with human-level labeling performance while only requiring lightweight human annotations.

Early works on 3D annotation have investigated automatic annotation models with weak annotations, e.g., 2D bounding boxes~\cite{wei2021fgr,qi2018frustumpnet, MAPGen2022, MTrans2022}, center-clicks~\cite{meng2020weakly, meng2021towards}, or 2D segmentation masks~\cite{mccraith2021lifting2d,wilson20203d_for_free_hdmap}. They rely on foreground/background 3D segmentation to preserve high-quality 3D annotations. While effective, such complicated annotation designs have required multi-staged training models, additionally processing point clouds, and designing special 3D operators and objective functions. 3D annotation in such cascaded pipelines is prone to be influenced by the failure modes of the early stages, e.g., the first 3D segmentation is a performance bottleneck for the final boxes regression stage. 

\begin{figure}[t]
\centering
\includegraphics[width=\columnwidth]{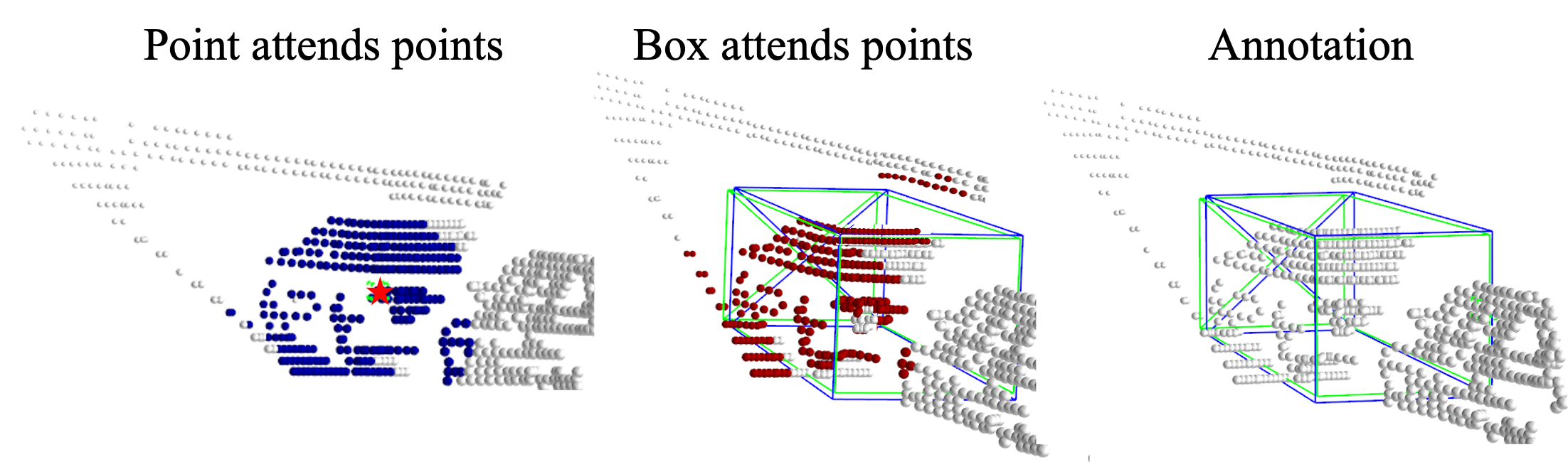} 
\caption{Attention maps directly from a CAT layer. We compute the attention from the reference point (red star) to all points of an object and display the points with the top 500 highest attention scores in blue. The attention groups the points within an object. The attention between the box and the points is displayed in red to describe the correlations between the box (in blue) and the points. Verge points in the background can also provide positive information about the box size and location, presumably making it easier to generate the 3D bounding box. Ground truth in green.}
\label{fig:attention}
\end{figure}

\begin{figure*}[t]
\centering
\includegraphics[width=0.93\textwidth]{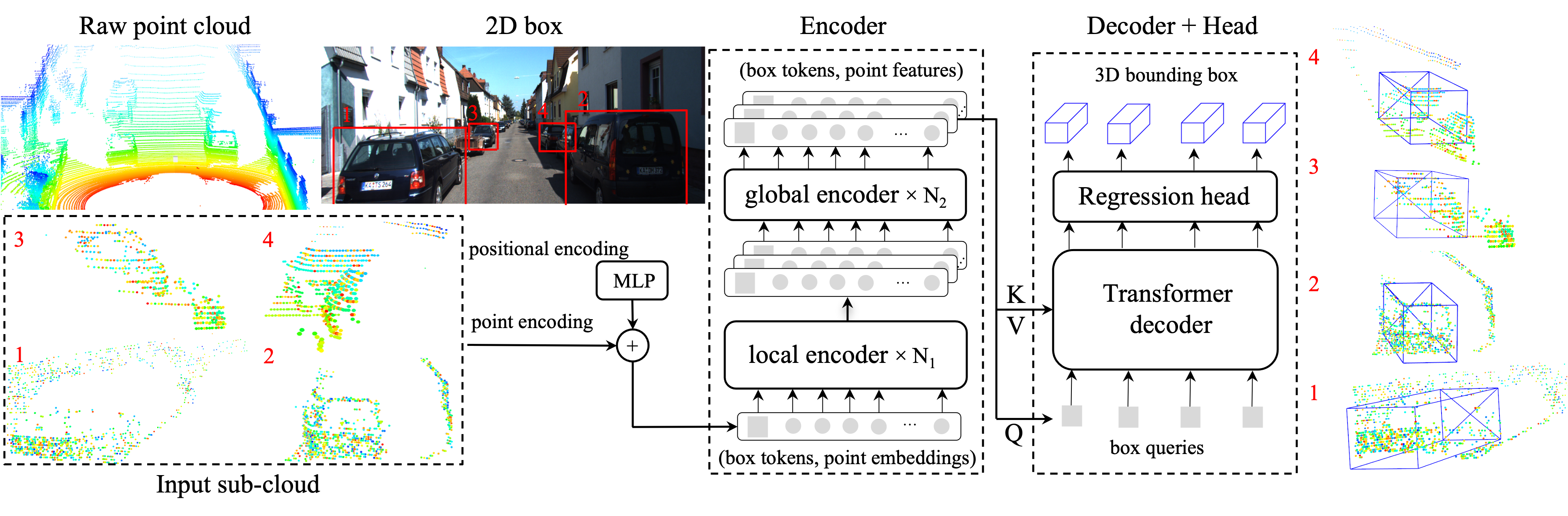} 
\caption{CAT is an end-to-end trainable context-aware Transformer-based 3D automatic annotator. It takes the frustum sub-clouds within 2D boxes as inputs and generates 3D bounding boxes.  The local encoder with $N_1$ layers explicitly models long-range contexts among all pairs of points at an object level, and the global encoder captures contextual interactions across objects (e.g., car 1, 2, 3, and 4) for point feature learning. A Transformer decoder then takes as input a small fixed number of box tokens corresponding to 3D box representations, which we call box queries ($Q$), and attends to point features ($K$ and $V$) from the encoder output. We pass the decoder results to regression heads (i.e., MLPs) that regress 3D bounding boxes.}
\label{fig:cat}
\end{figure*}

Acquiring high-quality 3D annotations is challenging since 3D point clouds are irregular, sparse, and unstructured, especially on hard samples with few points. To this end, multi-modality object detectors take advantage of images and point clouds, i.e., extracting features from each modality and fusing them for comprehensive predictions at the final stage for sparse point cloud tasks ~\cite{chen2017mv3d,huang2020epnet,zhao2021ACMNet, MTrans2022, MAPGen2022}. Although they have demonstrated a remarkable performance on the KITTI benchmark, they tend to increase the model complexity and calibration errors to combine multi-modal information. Meanwhile, only 5\% of camera information will be matched to the LiDAR point in the camera-to-LiDAR projection~\cite{liu2022bevfusion}. The projection of LiDAR-to-camera will drop the full-depth information in 3D point clouds.
Such density differences will become more drastic for sparser point clouds since projections between the camera and LiDAR in the multi-modality information combination are either semantically lossy or geometrically lossy~\cite{liu2022bevfusion}. Fortunately, we observe that using inter-object feature relations can conduct effective feature learning on point clouds by self-attention and avoid the semantic and geometric losses from multi-modality fusion. For example, hard samples with sparser point clouds can interact with others through inter-object relations and refine object-level features instead of using image features to explicitly complete point clouds. Inter-object feature relations can facilitate the samples' interactions, which helps annotate hard samples in heavily truncated, occluded, or very far instances.

However, such inter-object relations have been largely overlooked by previous research. In this paper, we leverage inter-object relations for 3D automatic annotation.
Self-attention in Transformer is designed to be permutation-invariant and encompass the dense token-wise relations that are especially relevant to extracting these inter- and intra-object relations. We argue that Transformer can solely rely on 3D point clouds to capture local dependencies among all-pairs points at an object level (intra-object) and global context-aware representations among objects at a scene level (inter-object) for 3D annotation.

To that end, we develop an end-to-end Context-Aware-Transformer (CAT), a simple-to-implement 3D automatic annotation method with a neat annotation workflow. We adopt the general encoder-decoder architecture. For the encoder, we make an essential change to adapt it to extract local and global contexts, namely, an intra-object encoder (local) and an inter-object encoder (global). As shown in Figure~\ref{fig:attention}, the local encoder allows each point to establish connections within an instance. Moreover, the background can positively impact the generated box size and location, e.g., the generated box should not enclose background points. Background points can help CAT determine the bounds of generated boxes. The global encoder provides scene contexts and feature relations between different objects. This way, hard samples with inter-object relations can capture the semantic patterns and implicitly refine representations instead of explicit point generation.

CAT removes many cumbersome designs, i.e., 3D segmentation, cylindrical object proposals generation, point cloud completion, multi-modality information combination, and complicated loss functions, while being simple to implement and understand. Unlike~\cite{MAPGen2022}, CAT does not need a ConvNet backbone on image features, and solely relies on 3D point clouds and Transformers trained from scratch. CAT is not only a conceptual recipe for 3D annotations but also allows us to perform comparatively on the KITTI dataset. In fact, CAT is empowered by its simplicity and elegance, and sets the new state-of-the-art on the KITTI benchmark.
We envision this work will serve as a simple yet effective baseline for future 3D autonomous annotation and inspire more research on rethinking more straightforward annotation methods for 3D. 

In summary, this paper makes the following contributions:
\begin{enumerate}
    \item We propose a simple yet effective end-to-end context-aware Transformer, CAT, that serves as an automated 3D-box annotator and decreases the heavy 3D human annotation burden. 
    \item We introduce the local and global encoders to extract intra-object and inter-object contextual information, performing self-attention along the sequence and batch dimensions, respectively. CAT can be local and global context-aware for hard samples with very few points to interact with each other and enhance 3D annotation.

    \item We show that CAT achieves new state-of-the-art performance compared to existing annotation methods on the KITTI benchmark, even without 3D segmentation, cylindrical object proposals generation or point cloud completion, and multi-modality information combination. 
\end{enumerate}

\section{Related Work}
\label{sec:related work}
We build upon prior works in 3D point cloud representation learning and 3D object annotations from weak annotations.

\subsection{Point Cloud Representation Learning}
Unlike 2D images, where pixels are arranged on regular grids and can be processed by CNNs, 3D point clouds are inherently sparse, unstructured, and scattered in 3D space: they are essentially set. These properties make typically convolutional architectures hardly applicable. 
Two ad hoc solutions to consume point clouds are transforming irregular points into voxels or grids and directly ingesting point clouds. Voxel-based networks~\cite{7353481_VoxNet,wu20153d,xie2018learning,yang2018foldingnet} divide point clouds into volumetric grids and employ 2D/3D CNNs for predictions.  Point-based methods~\cite{qi2017pointnet,qi2017pointnet++,lang2019pointpillars,shi2019pointrcnn,qi2018frustumpnet} directly consume raw point clouds to preserve more original geometric information for 3D tasks. 

As an alternative, attention-based models have been recently used to represent 3D point clouds for 3D vision tasks, e.g., retrieval~\cite{zhang2019pcan}, indoor and outdoor segmentation~\cite{Zhao_2021_ICCV}, object classification~\cite{liu2020tanet}, and 3D object detection~\cite{misra2021end,Pan_2021_CVPR}, since Transformer and attention models have revolutionized natural language processing~\cite{devlin2018bert,wolf2020Transformers} and computer vision~\cite{vaswani2017attention,dosovitskiy2020vit,liu2021swinTransformer,carion2020dert}. In Point Transformer~\cite{Zhao_2021_ICCV}, they design a point Transformer layer with vector self-attention layers and use it for constructing an Unet-like architecture for 3D point cloud segmentation and classification. In PCT~\cite{guo2021pct}, they make several adjustments on Transformers for 3D and introduce offset-attention and neighbor embedding to make a point Transformer framework suitable for point cloud learning. Adopting a Transformer encoder-decoder architecture, \cite{yu2021pointr} proposes PoinTr to reformulate point cloud completion as a set-to-set translation problem. By taking advantage of the solid capability to process long-range token-wise interactions and enhance information communications, Transformers begin their journey in 3D tasks. 

However, existing Transformer-based models for 3D point clouds bring in handcrafted inductive biases (e.g., local feature aggregation, neighbor grouping, and embeddings) and usually overlook the inter-object relation for effective feature learning on 3D tasks. Although the above methods apply specific 3D properties to modify the Transformer, we push the limits of the standard Transformer for 3D intra-object and inter-object feature extractions.

\subsection{3D Object Annotation from Weak Labels}
Human point cloud annotation is laborious~\cite{meng2021towards,wei2021fgr,MTrans2022,MAPGen2022}, hindering the training of high-quality 3D object detectors on massive datasets~\cite{Geiger2012CVPR_kitti,caesar2020nuscenes,sun2020scalability_waymo}. 2D weak annotations are more easily-acquired and cheaper compared with 3D labels. 3D object annotation from weak labels raises ever-growing attention. To our knowledge, very few works attempt to research 3D automatic annotation models. Using 2D boxes, FGR~\cite{wei2021fgr} proposes a heuristic algorithm to first locate key points and edges in the sub-point-cloud frustum after a coarse 3D segmentation and then estimate the 3D box. SDF~\cite{zakharov2020sdf} adopts the predefined models to process 3D segmented point clouds to estimate 3D annotations from 2D bounding boxes. WS3D~\cite{meng2021towards,meng2020weakly} explores center clicks as weak annotations to generate cylindrical object proposals on bird's view (BEV) maps and refine them to get 3D labels at the second stage. MAP-Gen~\cite{MAPGen2022} designs a 3-stage automated 3D-box annotation workflow with 2D weak boxes: first foreground segmentation, point generation, and final 3D box generations. MTrans~\cite{MTrans2022} leverages both LiDAR scans and camera images to generate precise 3D labels from weak 2D bounding boxes with a multi-task design of segmentation, point generation, and box regression. 

While they have demonstrated state-of-the-art performance on the KITTI benchmark, such complicated annotation designs involve pipelined training models, intermediately processing points, designing special loss functions, and multi-modality information combinations. Compared to these methods, our model is an end-to-end Transformer-based 3D annotator ( no convolutional backbone) that can be trained from scratch. CAT has removed many cumbersome designs, i.e., 3D segmentation, point completion, and multi-modality information combination. CAT requires minimal modifications to the vanilla Transformer layer.

\section{Methodology}
\label{sec: CAT}
This section describes CAT, simplifications in 3D point cloud autonomous annotation, and objective function.

\subsection{Data Preparation}
Our goal is to generate 3D pseudo labels that can be used to train any off-the-shelf 3D object detector (e.g., PointRCNN). 
Given LiDAR point clouds and weak 2D boxes, we first extract the frustum area tightly fitted with 2D bounding boxes, following~\cite{qi2018frustumpnet,wei2021fgr, MTrans2022, MAPGen2022}. 
With a known LiDAR-to-Image calibration matrix, a 3D point cloud in the form of $(x,y,z)$ can be projected onto its 2D image plane $(h, w)$ by the mapping function $f_{cal}$.
Therefore, the 2D projected sub-cloud $\mathcal{P}_{2D} \in \mathbb{R}^{N\times 2}$ within a 2D box can be defined as:
\begin{equation}
    \mathcal{P}_{2D} = \{(h, w)\; |\; (h, w) = f_{cal}(x, y, z),(h, w) \in \mathcal{B} \}. 
\end{equation}
Therefore, the frustum sub-cloud  $\mathcal{P}_{F} \in \mathbb{R}^{N\times 3}$ within its 2D bounding box can be extracted as:
\begin{equation}
    \mathcal{P}_{F} = \{(x, y, z)\; |\; f_{cal}(x, y, z) \in \mathcal{P}_{2D}\}, 
\end{equation}
where $(x,y,z)$ is the point coordinate, and $\mathcal{B}$ is the 2D region in the 2D box. We use $N$ to represent input frustum sub-cloud size (i.e., the number of points).

\subsection{CAT: Context-Aware Transformer}
CAT takes as input $B$ frustum sub-clouds $\mathcal{P}_{F} \in \mathbb{R}^{N\times 3}$ within 2D boxes, and generates 3D labels in the form of 3D bounding boxes ${Box}_{3D} \in \mathbb{R}^{B\times 7}$, where batch size is denoted as $B$. An input point sub-cloud is an unordered set of points associated with 3-dimensional coordinates $(x,y,z)$. The number of points varies from individual to individual. Some are very large when LiDAR scans are complete, but some are small due to occlusion, truncation, limited sensor resolution, light reflection, \textit{etc}. We use the random sampling from~\cite{qi2018frustumpnet,wei2021fgr, MTrans2022, MAPGen2022} to keep the same size (i.e., $N=1024$ points) of each sample and project them to $d=512$ dimensional point embeddings through a multilayer perceptron (MLP). The resulting point embeddings are passed through the CAT encoder to obtain a set of features. The additional box tokens are taken as box representations along with point embeddings in the encoder, inspired by the class token in ViT~\cite{dosovitskiy2020vit}. A decoder takes as input box tokens as queries, points features as keys, and values from encoder output and predicts 3D bounding boxes.

Both encoder and decoder employ standard Transformer layers with the pre-norm, self-attention, and MLP. We refer the readers to~\cite{vaswani2017attention,carion2020dert} for more details on the Transformer encoder and decoder. Figure~\ref{fig:cat} illustrates our CAT model for 3D automatic annotation.

\subsection{Encoder for Point Cloud Representation}
CAT encoder comprises an intra-object encoder (local) and an inter-object encoder (global) to perform self-attention along the sequence and batch dimensions. The local encoder explicitly models local contexts among all-pairs points at an object level. The resulting features are then fed to the global encoder to capture the contextual interactions at a scene level for further point feature learning.

\subsubsection{Local Encoder}
The input projection step provides a set of $N$ features of $d=512$ dimensions using an MLP with two hidden layers as point embeddings. These 512-d point embeddings associated with 7 box tokens are then passed to an intra-object encoder  (local) to produce $N+7$ features of 512 dimensions. The 7 box tokens represent the 3D bounding box in the form of $(x,y,z, width, length, height, yaw)$. The local encoder applies $N_1$ Transformer layers composed of multiheaded self-attentions and non-linear channel MLPs. We use the standard dot-product self-attention formulation in~\cite{yu2021pointr} to extract the feature relations between all pairs of input points and box tokens. 

By leveraging messages passing among all pairs of points, all points of an object can be equally considered since relevant contextual information can be anywhere. Furthermore, the object's background points also participate in the message communication and provide positive information for the final box generation. Figure~\ref{fig:attention} shows the background points can contribute to positive supervision of the bounds of 3D-generated boxes.

\subsubsection{Global Encoder}
It is valuable to extract global contextual interactions and feature correlations between different instances for 3D annotations, which are commonly omitted in other annotation methods. The inter-object relationships are informative in describing the scenes of point clouds, especially for hard samples to attend to each other to obtain the feature patterns and refine 3D representations, enhancing the 3D automatic annotation.  

To facilitate the encoder to better communicate among different objects, we further devise an inter-object encoder (global) composed of $N_2$ layers to capture global interactions across objects along the batch ($B$) dimension, instead of the sequence ($N+7$) dimension in the local encoder. 
The local encoder provides local feature relations of $(box\_token, point\_feature)$ in the dimension of $B\times (N+7)\times d$ that is subsequently passed to the global encoder. Since the global encoder performs self-attentions in the batch $B$ dimension, local encoder output should be transposed to $(N+7)\times B\times d$ before passing it into the batch-wise global encoder. The global feature can communicate to all objects and their box tokens in each mini-batch. This way, each object can connect with all objects during training. 

With an inter-object relationships extraction, we can capture feature relations and promote message passing among objects for scene-level understanding. Specifically, global contextual interactions can be constructive for hard samples where inter-related objects are selected from attention weights to complete features of hard samples, further improving the quality of generated 3D boxes.

\begin{table*}[t]
    \centering
    \setlength{\tabcolsep}{3pt}
    \resizebox{\linewidth}{!}{
    \begin{tabular}{lccccccccc}
\toprule
\multirow{2}{*}{Method} & \multirow{2}{*}{Modality} & \multirow{2}{*}{Full Supervision} & \multicolumn{3}{c}{$\text{AP}_{3D}(IoU=0.7)$} & \multicolumn{3}{c}{$\text{AP}_{BEV}(IoU=0.7)$}\\

\cmidrule(lr){4-9}
& & & \multicolumn{1}{c}{Easy} & \multicolumn{1}{c}{Moderate} & \multicolumn{1}{c}{Hard} & \multicolumn{1}{c}{Easy} & \multicolumn{1}{c}{Moderate} & \multicolumn{1}{c}{Hard} \\

\midrule
PointRCNN~\cite{shi2019pointrcnn} & LiDAR & \cmark & \textit{86.96} & \textit{75.64} & \textit{70.70} & \textit{92.13} & \textit{87.39} & \textit{82.72} \\
MV3D\cite{chen2017mv3d} &LiDAR & \cmark & 74.97 & 63.63 & 54.00 & 86.62 & 78.93 & 69.80 \\
F-PointNet\cite{qi2018frustumpnet} & LiDAR& \cmark & 82.19 & 69.79 & 60.59 & 91.17 & 84.67 & 74.77 \\
AVOD\cite{ku2018AVOD} &LiDAR & \cmark & 83.07 & 71.76 & 65.73 & 90.99 & 84.82 & 79.62 \\
SECOND\cite{yan2018second} & LiDAR& \cmark & 83.34 & 72.55 & 65.82 & 89.39 & 83.77 & 78.59 \\
PointPillars\cite{lang2019pointpillars} &LiDAR & \cmark & 82.58 & 74.31 & 68.99 & 90.07 & 86.56 & 82.81 \\
SegVoxelNet\cite{yi2020segvoxelnet} &LiDAR & \cmark & 86.04 & 76.13 & 70.76 & 91.62 & 86.37 & 83.04 \\
Part-A$^2$\cite{shi2020partA2} & LiDAR& \cmark & 87.81 & 78.49 & 73.51 & 91.70 & 87.79 & 84.61 \\
PV-RCNN\cite{shi2020pvrcnn} &LiDAR & \cmark & 90.25 & 81.43 & 76.82 & 94.98 & 90.65 & 86.14 \\
\midrule

WS3D~\cite{meng2020weakly} & LiDAR & BEV Centroid & 80.15 & 69.64 & 63.71 & 90.11 & 84.02 & 76.97 \\
WS3D(2021)~\cite{meng2021towards} & LiDAR & BEV Centroid & 80.99 & 70.59 & 64.23 & 90.96 & 84.93 & 77.96 \\
FGR~\cite{wei2021fgr} & LiDAR & 2D Box & 80.26 & 68.47 & 61.57 & 90.64 & 82.67 & 75.46 \\

MAP-Gen~\cite{MAPGen2022}      & LiDAR+RGB & 2D Box & 81.51 & 74.14 & 67.55 & 90.61 & 85.91 & 80.58\\
MTrans~\cite{MTrans2022}       & LiDAR+RGB & 2D Box & 83.42 & 75.07 & 68.26 & 91.42 & 85.96 & 78.82\\
\midrule
CAT (ours)   & LiDAR & 2D Box & \textbf{84.84} & \textbf{75.22} & \textbf{70.05} & \textbf{91.48} & \textbf{85.97} & \textbf{80.93} \\
\bottomrule
    \end{tabular}
    }
    \caption{Results of KITTI official \emph{test} set, compared to the fully supervised PointRCNN and other weakly supervised baselines.}
    \label{tab1:compare_weak_methods_KITTItest}
\end{table*}


\subsection{Decoder for 3D Bounding Box Generation}
To generate high-quality 3D bounding boxes, CAT should focus on essential parts of point features for 3D box regression. We introduce a decoder for the box to one-way attends point features to refine object-level features for the final box regression. We take box tokens as queries and point features as contexts from global encoder output $(box\_tokens, point\_features)$ to query point features.

Following~\cite{carion2020dert}, our decoder operates $B$ objects in parallel using multiheaded self- and cross-attention mechanisms in one batch $B$.
Box queries are learned $box\_tokens$ from global encoder output. Each query has 7 box tokens in the form of $(x,y,z, width, length, height, yaw)$.
The decoder takes the $point\_features$ as keys and values $(K, V)$ and $box\_tokens$ as queries $(Q)$. The resulting decoder outputs are then used to generate boxes, ${Box}_{3D} \in \mathbb{R}^{B\times 7}$. 
In our framework, box queries represent 3D boxes in 3D space around which our final 3D boxes in $B\times 7$ are generated by three MLP heads to predict locations in $(x,y,z)$, dimensions in $(width, length, height)$ and rotation along the z-axis $yaw$, respectively. Each MLP head consists of one linear layer and leaky ReLU nonlinearity. Thus, the Transformer decoder in~\cite{carion2020dert} has not many specific modifications for 3D. 

Using self- and cross-attention over the global encoder output, the CAT decoder globally refines $B$ object-level features for 3D box regression, while being able to use $B$ objects' global point features as contexts.

\subsection{Loss Function}
To train the model, we use a loss directly on the dIoU as in~\cite{zheng2020distanceiou} for each generated box and its matched ground truth box denoted as $\mathcal{L}_{box}$. Moreover, we introduce a direction loss $\mathcal{L}_{dir}$ for a binary direction classification, since the IoU metric is direction-invariant~\cite{MTrans2022}. The direction classification is performed by an MLP where orientation within $[-\pi/2, \pi/2)$ is predicted as the front, and $[\pi/2, \pi] \cup [-\pi, -\pi/2)$ as the back. This encourages our model to learn the right box direction, a property that helps our regression on yaws. We use Cross-Entropy loss for the direction loss $\mathcal{L}_{dir}$.
\begin{equation}
    \mathcal{L} = \lambda_{box}\mathcal{L}_{box} + \mathcal{L}_{dir},
\end{equation}
where the $\lambda_{box}$ is a weight, we empirically set it as 5.

Our loss function is a weighted sum of two terms above. CAT removes many handcrafted designs on the loss function while being way simpler to implement and understand.

\subsection{Position Encoding}
Standard positional encoding for vision tasks is crafted manually, e.g., sinusoidal position embedding~\cite{vaswani2017attention} based on sine and cosine functions and normalized range values~\cite{vaswani2017attention}. In 3D point cloud processing, 3D points contain information about $(x,y,z)$ coordinates that are natural positional information. For brevity, we go beyond this by introducing an MLP for trainable and parameterized positional embeddings. Trainable positional embedding on 3D coordinates is essential to adapt the 3D structure for point clouds. Notably, our model's more straightforward, parameterized, trainable positional embedding outperforms other position encoding schemes. An MLP is used to encode the 3D position $pos$ as:
\begin{equation}
    pos =  MLP (x,y,z).
\end{equation}
We concatenate box tokens and points in our sequence, $concat(box\_tokens, pos)$. For dimension consistency, we manipulate another MLP with one linear layer in $pos$ to describe the position of each entity in the sequence.

\begin{table*}[ht]
    \centering
    \setlength{\tabcolsep}{12pt}
    \resizebox{\linewidth}{!}{
    \begin{tabular}{lcccccc}
\toprule
\multirow{2}{*}{Method} & \multirow{2}{*}{Modality} & \multirow{2}{*}{Full Supervision} & \multicolumn{3}{c}{$\text{AP}_{3D}(IoU=0.7)$} \\
\cmidrule(lr){4-6}
& & & \multicolumn{1}{c}{Easy} & \multicolumn{1}{c}{Moderate} & \multicolumn{1}{c}{Hard} \\

\midrule
PointRCNN~\cite{shi2019pointrcnn} &LiDAR & \cmark & 88.88 & 78.63 & 77.38 \\

\midrule

WS3D\cite{meng2020weakly} & LiDAR & BEV Centroid & 84.04 & 75.10 & 73.29 \\
WS3D(2021)\cite{meng2021towards} & LiDAR & BEV Centroid & 85.04 & \textit{75.94} & \textit{74.38}\\
FGR\cite{wei2021fgr} & LiDAR & 2D Box & \textit{86.68} & 73.55 & 67.91\\

MAP-Gen~\cite{MAPGen2022}      & LiDAR+RGB & 2D Box & 87.87 & 77.98 & 76.18\\
MTrans~\cite{MTrans2022}       & LiDAR+RGB & 2D Box & 88.72 & 78.84 & 77.43\\
\midrule
CAT (ours)   & LiDAR & 2D Box & \textbf{89.19} & \textbf{79.02} & \textbf{77.74}\\
\bottomrule
    \end{tabular}
    }
    
    \caption{Results of KITTI \emph{val} set, compared to the fully supervised PointRCNN and other weakly supervised baselines.}
    \label{tab2:compare_weak_methods_KITTIval}
\end{table*}

\section{Experiments}
\label{sec: experiments}
\paragraph{Dataset and metrics.}
We adopt the KITTI Benchmark~\cite{Geiger2012CVPR_kitti} for CAT evaluation. The KITTI dataset is one of the best-known benchmarks for 3D detection in autonomous driving~\cite{Geiger2012CVPR_kitti}. The 3D detection benchmark contains 3712 frames of data with 15,654 vehicle instances on the \emph{train} split and 3769 frames on the \emph{val} split. For a fair comparison, we used the official split with 500 frames for training and 3769 for evaluation. We use the same dataset pre-process to extract the objects within 2D bounding boxes in~\cite{wei2021fgr, MTrans2022, MAPGen2022}. We follow the official evaluation protocol provided by KITTI. Detection outcomes of CAT-trained PointRCNN are evaluated on three standard tasks: Easy, Moderate, and Hard Tasks. We report the detection performance using the 3D mean Average Precision (mAP) threshold of 0.7 mean Intersection over Union (mIoU).

\paragraph{Implementation details.}

We implement CAT using PyTorch~\cite{paszke2019pytorch} and employ standard Transformer layers to implement the CAT encoder and decoder. The local encoder has $N_1=8$ layers, each using a multiheaded self-attention with eight heads and an MLP with two linear layers and one ReLU nonlinearity. The global encoder with $N_2=3$ layers closely follows the local encoder settings except that it is implemented to perform self-attention along the batch dimension. The decoder has three Transformer decoder layers composed of multiheaded self-attentions, cross-attentions, and MLPs. The prediction heads for box regression are two-layer MLPs with a hidden size of 1024. 
CAT is optimized using the Adam optimizer with the learning rate of $10^{-4}$ decayed by a cosine annealing learning rate scheduler and a weight decay of 0.05. We train CAT on a single RTX 3090 with a batch size of 24 for 1000 epochs. We use standard data augmentations of random shifting, scaling, and flipping. Following~\cite{MTrans2022, MAPGen2022,wei2021fgr,qi2018frustumpnet}, we filter the objects with at least 30 total points and 5 foreground points for CAT training since it is not feasible for 3D annotation on some hard samples with no foreground LiDAR points. At inference time, our end-to-end labeling system runs near real-time, with a latency of 28 milliseconds per object, which can be 3-5 times faster than pipelined annotation models. We use the 3D detector of PointRCNN for 3D object detection.

\subsection{CAT on 3D Automatic Annotation}
We now evaluate the performance of CAT as an auto labeler on the KITTI dataset. CAT is trained with 500 frames of labeled 3D data to generate 3D pseudo labels from weak 2D boxes of the KITTI dataset. CAT-generated 3D pseudo labels are used to train the 3D object detector of PointRCNN. We compare it to other baselines on the KITTI \emph{test} and \emph{val} sets, as shown in Table~\ref{tab1:compare_weak_methods_KITTItest} and Table~\ref{tab2:compare_weak_methods_KITTIval}.

\subsubsection{Observations}

On the \emph{test} set (Table~\ref{tab1:compare_weak_methods_KITTItest}), CAT achieves new state-of-the-art performance, even though we remove 3D segmentation, point cloud completion, and multi-modality information combination, compared to baselines~\cite{wei2021fgr,meng2021towards, meng2020weakly, MTrans2022, MAPGen2022}. Only requiring 500 frames of labeled data, CAT-trained PointRCNN can yield 98.63\% of the original model supervised with full human annotations. Compared to the latest multi-modal MTrans~\cite{MTrans2022}, CAT-trained PointRCNN improves the 3D detection accuracy on $AP_{3D}$ by 1.42\%, 0.15\%, and 1.79\% on Easy, Moderate, and Hard tasks, respectively. The hard task usually suffers more from irregularity and sparsity issues, since hard samples are usually much sparser than easy samples. The most significant improvement of 1.79\% $AP_{3D}$ on the Hard Task benefits from inter-object feature relations learning on hard samples. In CAT, the sparsity issue can be mitigated by the scene contexts and feature correlations passing between objects. 

As shown in Table~\ref{tab2:compare_weak_methods_KITTIval}, PointRCNN trained with CAT-generated pseudo labels yields on-par performance of the original model under the supervision of ground truth 3D labels on the KITTI \emph{val} split. CAT-trained PointRCNN outperforms all existing annotation methods on the KITTI \emph{val} set. Experiments on the KITTI \emph{val} split show consistent improvements over state-of-the-art models of $AP_{3D}$ while simultaneously simplifying the automatic annotation workflow and designs of the model architecture and loss function.

The experiment results validate that a Transformer-based model is competitive with state-of-the-art baselines tailored for 3D automatic annotation. Our CAT can be a simple yet effective automated 3D annotator for high-quality box generation, decreasing the heavy human annotation burden. The observation aligns with the motivation of applying Transformer-based models to capture the local and global contexts and feature relations of irregular and sparse point clouds for 3D representation learning. CAT can generate high-quality 3D labels using intra- and inter-object encoders and a decoder with a streamlined annotation workflow.

\subsubsection{Qualitative Results}

\begin{figure*}[ht]
\centering
\includegraphics[width=1.9\columnwidth]{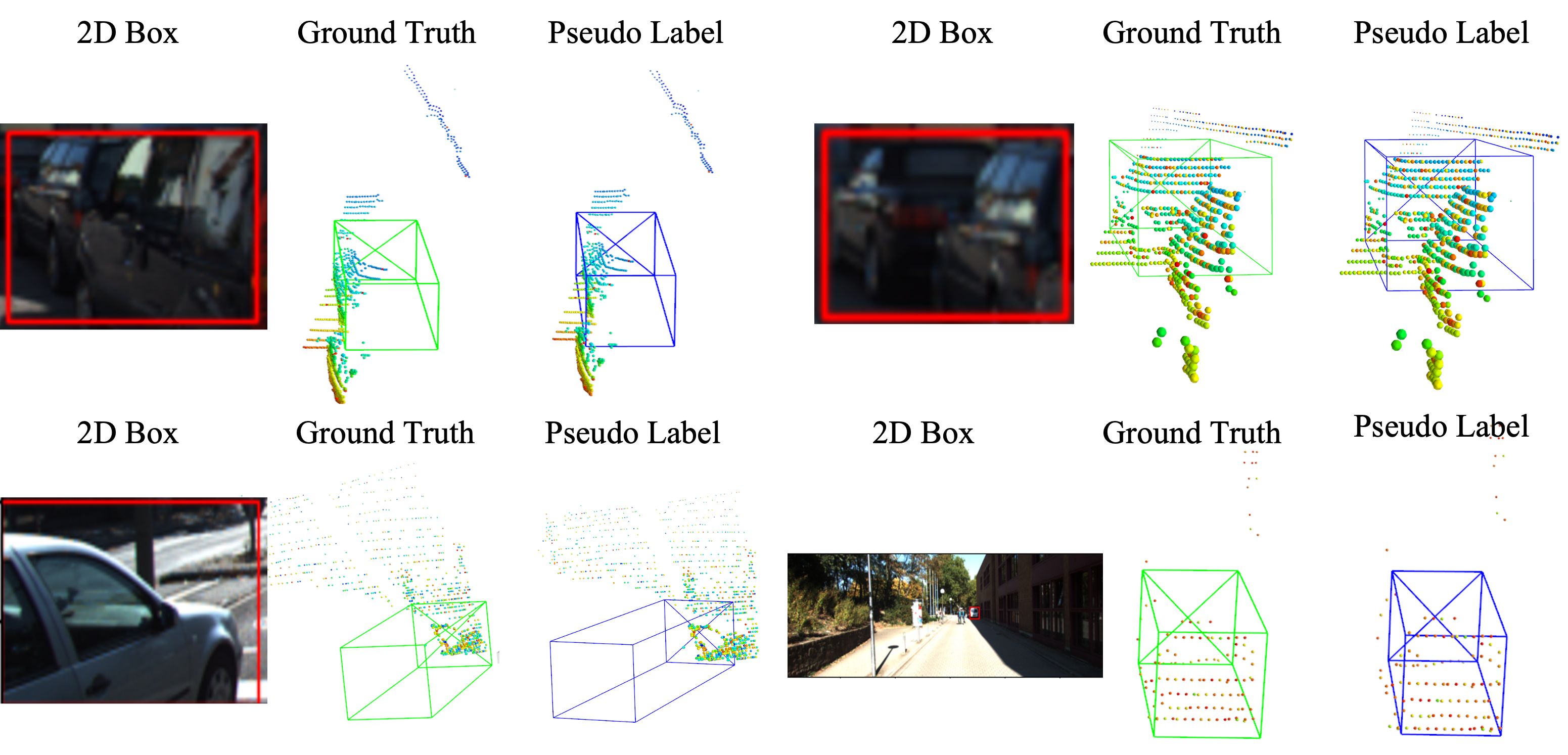} 
\caption{Qualitative annotation results using CAT on the KITTI \emph{val} split. CAT does not use the image information (here for visualization) and generates 3D bounding boxes from point clouds. For hard samples with heavily truncated (bottom left), occluded (top left), and very far instances (bottom right), CAT can generate an amodal box, e.g., the full extent of cars.}
\label{fig:qualitative results}
\end{figure*}

In Figure~\ref{fig:qualitative results}, we visualize a few CAT-generated pseudo labels and ground truth boxes on hard samples with few points. CAT can generate high-quality bounding boxes for hard samples in heavily truncated or occluded, or very far instances, by attending to other easy samples (not highly occluded) and using inter-object relations to refine 3D representations. 
These qualitative results demonstrate the effectiveness of CAT and also verify that it is possible to combine the semantic information (from foregrounds and backgrounds) within 2D boxes and the 3D geometric structure of point clouds for 3D automatic annotations. 

\begin{table}[ht]
    \centering
    \setlength{\tabcolsep}{1.5pt}
    \resizebox{0.48\textwidth}{!}{
    \begin{tabular}{lccccccccc}
    \toprule 
    \multirow{2}{*}{} & \multirow{2}{*}{Local} & \multirow{2}{*}{Global} & \multirow{2}{*}{Decoder} & \multirow{2}{*}{Pos. }  & \multicolumn{4}{c}{Metric} \\
    \cmidrule(lr){6-9} &Enc. &Enc. & & Embed. &  mIoU & Recall & mAP & $\text{mAP}_{R40}$\\
    \midrule
    Model A & \cmark & &  &  &  65.01 & 54.50 & 56.31  & 59.24 \\
    \midrule
    Model B & \cmark & \cmark & &  &  70.70 & 65.21 & 83.20 & 85.89   \\
    Model C & \cmark & \cmark & \cmark & &  72.88 & 70.84 & 85.83 & 87.72  \\
    Model D & \cmark & \cmark & \cmark & \cmark (Sine)  & 72.13  & 69.66  & 85.21  & 87.34  \\

    \midrule
    Ours(500) & \cmark & \cmark & \cmark & \cmark (MLP) & \textbf{73.71} & \textbf{72.78} & \textbf{87.04} & \textbf{89.80} \\ 
    Ours(all) & \cmark & \cmark & \cmark & \cmark (MLP) & 78.91 & 84.13 & 91.30  & 93.38  \\
    \bottomrule
    \end{tabular}
    }
    \caption{Ablation Results.}
    \label{tab:ablation}   
\end{table}

\subsection{Ablations}
We conduct a series of experiments to confirm the effectiveness of each module in our proposed model, namely, the local encoder, global decoder, and decoder. Meanwhile, we also study the choice of position embedding strategies for CAT training. The 3D pseudo labels are compared with human annotations on the metrics of mIoU, recall with IoU threshold of 0.7, mAP, and mAP$_{R40}$ over the Car category. 

In Table~\ref{tab:ablation}, we regard the standard Transformer encoder (Local Enc.) as a baseline. 
Model A (Local Enc.) shows the local encoder extracts intra-object interactions among points.
Model B (Local+Global Enc.) brings a performance boost of 5.6\%, 10.7\%, 26.8\%, 26.6\% on mIou, Recall, mAP, and mAP$_{R40}$, demonstrating that global encoder models inter-object relations and facilitates samples' interactions, particularly helpful on hard samples.
Model C (Local+Global Enc.+Decoder) further improves mIoU by up to 2.31\%, clarifying that the decoder helps to refine representations for more precise box regression.
Model D (Local+Global Enc.+Decoder+Pos Embed.) achieves the final 72.78\% Recall and 87.04\% mAP in an MLP embedding way, verifying positional encoding on 3D coordinates is essential to adapt to the 3D structure. We can observe that the more straightforward parameterized and trainable positional embedding of MLP outperforms other hand-designed position encoding schemes in our model.
Meanwhile, we evaluate the effectiveness of box tokens in CAT. Seven Box tokens represent seven box elements (XYZ localization, HWL size, and yaw orientation), as box representations.
Removing box tokens in the CAT encoder results in a 2.69\% IoU drop, suggesting queries from box tokens instead of position encoding in~\cite{carion2020dert} can help to refine object-level features for the final 3D box generation in the decoder. 


\section{Conclusion}
We present CAT, a simple yet effective end-to-end Transformer model for 3D automatic annotation. CAT directly consumes LiDAR point clouds as an auto labeler to generate 3D bounding boxes from weak 2D boxes. CAT requires only a few handcrafted designs tailored for 3D autonomous annotation and effectively reduces human annotation burdens. We adopt the general encoder-decoder architecture, where the CAT encoder consists of an intra-object encoder (local) and an inter-object encoder (global). The local encoder extracts local dependencies among all-pairs points at an object level. The global one captures global contextual interactions and scene contexts at a scene level for 3D representation learning. The decoder refines feature relations between box representations and point features for 3D box generation. We show that using both local and global encoders is critical for high-quality 3D box generation. Endowing such local and global contextual communications to sparse point clouds, CAT outperforms existing state-of-the-art baselines on the KITTI dataset with a streamlined 3D annotation workflow. We hope this work will serve as a simple baseline to inspire more research on a simpler annotation model for future 3D automatic annotation tasks.

This approach for 3D annotation also comes with new challenges, particularly regarding annotating small objects. Current 3D automatic annotation models require years of careful designs to handle similar issues, and we will further research this for CAT.

\section{Acknowledgments}
This work is supported in part by the Research Grants Council of the Hong Kong Special Administrative Region, China, under the General Research Fund (GRF) projects 17206020 and 17209721.



\bibliography{ref}

\end{document}